\documentclass[journal]{IEEEtran}

\usepackage[T1]{fontenc}
\usepackage[utf8]{inputenc}

\usepackage[style=ieee,backref=true,isbn=false,eprint=false,doi=false]{biblatex}
\usepackage{flushend}

\ifCLASSINFOpdf
  \usepackage[pdftex]{graphicx}
  \graphicspath{{./fig/}}
  \DeclareGraphicsExtensions{.eps,.pdf,.jpeg,.jpg,.png}
\else
\fi

\usepackage{epstopdf}

\usepackage[table]{xcolor} 
\usepackage{multirow}
\definecolor{g1}{gray}{0.9}
\definecolor{g2}{gray}{0.85}
\usepackage{booktabs}

\usepackage{amsmath,amssymb}
\usepackage{array}

\ifCLASSOPTIONcompsoc
  \usepackage[caption=false,font=normalsize,labelfont=sf,textfont=sf]{subfig}
\else
  \usepackage[caption=false,font=footnotesize]{subfig}
\fi

\usepackage{dblfloatfix}

\usepackage{url}

\usepackage[colorlinks=true]{hyperref}
\usepackage{blindtext}
\usepackage[noabbrev,nameinlink]{cleveref}
\crefname{subsection}{subsection}{subsections}

\usepackage{logicpuzzle}
\usepackage{makecell}
\usepackage{float}
\usepackage{afterpage}

\newcommand{\added}[1]{#1}  
\newcommand{\removed}[1]{}
\newcommand{\changed}[2]{\added{#2}}
\newcommand{\mc}[3]{\multicolumn{#1}{#2}{#3}}
\newcommand{\mr}[3]{\multirow{#1}{#2}{#3}}
\newcommand{\nop}[1]{}

\newcommand{\myfigscale}[0]{0.91}

\begin{document}
   \title{Multi-Armed Bandits for Minesweeper:\\Profiting from Exploration-Exploitation Synergy}

   \author{
      Igor~Q.~Lordeiro, Diego B. Haddad,~\emph{IEEE member}, and~Douglas~O.~Cardoso%
      \thanks{
         Igor Q. Lordeiro (corresponding author), Diego B. Haddad and Douglas O. Cardoso are with the Department of Computer Engineering, Federal Center for Technological Education Celso Suckow da Fonseca (CEFET-RJ), Petrópolis, RJ, Brazil.
         E-mails: \href{mailto:igor.lordeiro@aluno.cefet-rj.br}{igor.lordeiro@aluno.cefet-rj.br}, \href{mailto:diego.haddad@cefet-rj.br}{diego.haddad@cefet-rj.br} \href{mailto:douglas.cardoso@cefet-rj.br}{douglas.cardoso@cefet-rj.br}.
      }
   }

   %


   \maketitle

   \begin{abstract}
      A popular computer puzzle, the game of Minesweeper requires its human players to have a mix of both luck and strategy to succeed.
      Analyzing these aspects more formally, in our research we assessed the feasibility of a novel methodology based on Reinforcement Learning as an adequate approach to tackle the problem presented by this game.
      For this purpose we employed Multi-Armed Bandit algorithms which were carefully adapted in order to enable their use to define autonomous computational players, targeting to make the best use of some game peculiarities.
      After experimental evaluation, results showed that this approach was indeed successful, especially in smaller game boards, such as the standard beginner level.
      Despite this fact the main contribution of this work is a detailed examination of Minesweeper from a learning perspective, which led to various original insights which are thoroughly discussed.
   \end{abstract}

   \begin{IEEEkeywords}
      Computer Games, Reinforcement Learning, Greedy Policy, UCB, Transfer Learning
   \end{IEEEkeywords}

   \IEEEpeerreviewmaketitle

   \section{Introduction \label{sec:intro}}
      \IEEEPARstart{M}{inesweeper} is a single-player computer puzzle game originating from the 1960s.
      Since its initial release it has evolved in diverse aspects, having many different iterations and being included in various operating systems, most noticeably in members of the Microsoft Windows family.
      Its small set of rules as well as straightforward gameplay mechanics boosted its popularity up to the point of being considered one of the most successful games ever~\cite{cobbett_2009}.

      This game has been analyzed from a perspective of theory of computation, and its solution was related to NP-completeness~\cite{Kaye2000} and co-NP-completeness~\cite{Scott2011} according to the premises considered.
      Such results support the notion that the game is indeed challenging despite its apparent simplicity.
      Reinforcement Learning (RL)~\cite{sutton2018reinforcement,2010Szepesvari} has been used to tackle general NP-hard problems~\cite{YangJLSZ18,BelloPL0B17,Kenshin2019} as well as to create AI players for classic games such as Battleships, Chess, Shogi and Go~\cite{Ladislav2013,Silver1140}.
      The same goes for Minesweeper, which was approached using RL following some general, frequently used guidelines in this sense~\cite{Olivier2013}.
      This work is aimed at better exploiting some game properties previously overlooked, which led to a novel RL-oriented modeling whose exploration provided interesting insights and results.

      Using Multi-Armed Bandit algorithms~\cite{Slivkins19} and without the aid of any heuristics to pursue this goal, our agents managed to achieve win rates of over $75 \%$ and $55 \%$ in the beginner and intermediate difficulty levels, respectively. 
      Though we expected for a purely greedy agent to be unrivaled in the setting enforced by the game of Minesweeper, UCB agents were able to achieve convincing results, only faltering in higher levels.
      While overcoming the UCB agent, the greedy approach also managed to challenge an initial hypothesis of correspondence between knowledge extent and performance in this scenario, as despite its higher win rate, its number of actions learned is significantly lower.

      The remainder of this paper is organized as follows.
      \Cref{sec:walk} presents the game of Minesweeper, from its basic concepts up to its mechanics and strategic principles.
      A multifaceted overview of related research can be found in \cref{sec:related}.
      \Cref{sec:method} renders the original ideas we conceived as well as a thorough discussion of them.
      Practical results obtained from experimenting on the aforementioned ideas are presented in \cref{sec:exp}, which also explains how performance was assessed.
      At last, \cref{sec:end} provides some closing remarks and indicates possible continuations of this work.

   \section{A Brief Game Walk-Through \label{sec:walk}}
      A Minesweeper game is played using an $m \times n$ board of tiles.
      There are $k$ hidden mines which are randomly placed just after the always-safe first play/click.
      Considering the classic Windows 98 version of the game, difficulty levels vary between beginner ($8 \times 8$ board, 10 mines), intermediate ($16 \times 16$ board, 40 mines) and expert ($16 \times 30$ board, 99 mines).
      Besides these presets the game can be played in a custom format:
      the notation $r \times c \times b$ is used from now on in this paper to represent a $r \times c$ board with $b$ mines.
      The player's task is to uncover all tiles that do not feature a mine.

      A short description of the game rules~\cite{wikipedia_minesweeper} follows.
      As the player clicks on a tile, a few things can happen:
      if a mine is uncovered, game over;
      otherwise, the tile then becomes exposed, showing the number of mines in its eight neighboring tiles;
      if such number is zero, then the tile becomes blank and all the surrounding tiles become exposed, recursively.
      It is also possible to place a flag on a tile which is assumed to feature a mine, acting as a marker to prevent clicking on the said tile afterwards.
      On the other hand, the player can regret placing any flag and remove it.

      In a broad sense, rational playing relies on information provided by numbered tiles to deduce where mines are hidden and which tiles are safe.
      \Cref{fig:game_example} exemplifies a game state and the analysis which could be used to decide the next moves.
      While some game situations are inherently ambiguous, most of them can be solved based on similar reasoning.
      Occasionally this strategy needs to be complemented with conjecturing where mines are located and consequently finding contradictions.
      This clearly suggests to approach the game as a Constraint Satisfaction Problem (CSP), and indeed diverse works in the literature do likewise in some regard~\cite{TangJH2018, Olivier2013}.

      \begin{figure}[ht]
         \centering
         \begin{logicpuzzle}[color=gray!40, rows=5, columns=5, scale=0.5, width=0.296\columnwidth]
            \framepuzzle

            \fillcell{1}{5}
            \fillcell{2}{5}
            \fillcell{3}{5}
            \fillcell{4}{5}
            \fillcell{5}{5}
            \fillcell{1}{4}
            \fillcell{5}{4}

            \setrow{3}{1, 1, {}, 1, 1}
            \setrow{4}{F, 3, 1, 2, F}
         \end{logicpuzzle}
      
         \caption{
            An ongoing Minesweeper game in a $5\times5$ board with 4 mines.
            A pair of mines, each marked by a flag (F), are trivial to detect thanks to the 1-tiles beside them.
            It is impossible to find the last 2 mines from the information of a single tile, but it can be inferred that they are located in the 3 leftmost tiles of the first row, according to the 3-tile on position $(2,2)$.
            Therefore the last two tiles of the first row are safe, and clicking on them would lead to solving this game.
            \label{fig:game_example}
         }
      \end{figure}

   \section{Related Work \label{sec:related}}
      Reinforcement learning is a popular technique for creating agents to play games.
      Its first application in this regard dates back to 1959 with \Textcite{Samuel1959} creating a Checkers agent.
      More recently, RL has played a defining role on the development of competitive agents for games, with \Textcite{Ladislav2013} applying it to Battleships.
      More notably, \Textcite{Silver1140} were able to achieve groundbreaking results in Chess, Shogi and Go with AlphaZero. 
      Recent advances also include the work of \Textcite{Jaskowski18} in the game of 2048, \Textcite{McPartlandG11} and \Textcite{GlavinM15} in First Person Shooter (FPS) games and \Textcite{PintoC19} in Fighting Games.
      
      Minesweeper automatic solving was first approached by \Textcite{ADAMATZKY1997127} in 1997, who devised a cellular automaton to play it.
      However, the game was studied from a computational complexity perspective only in 2000, when \textcite{Kaye2000} proved that the problem of assigning mines to covered tiles, which was called Minesweeper consistency problem, is NP-complete.
      \Textcite{Scott2011} came to the conclusion that finding whether or not there is at least one covered tile whose mine status could undoubtedly be decided in any game instance, called Minesweeper inference problem, is co-NP-complete.
      
      Following research had its focus mainly on constraint-satisfaction and heuristic methods.
      \Textcite{Olivier2013} used a hybrid of Upper Confidence Trees and Heuristic CSP to achieve win rates of 80.2\%, 74.4\%, and 38.7\% in the beginner, intermediate and expert levels, respectively.
      Meanwhile \Textcite{Jinzheng2017} obtained win rates of 81.6\%, 78.1\%, and 39.6\% respectively for the beginner, intermediate and expert levels combining multiple heuristics.
      
      Regarding our approach to Minesweeper, \Textcite{sutton2018reinforcement,Slivkins19} didactically introduce the Multi-Armed Bandit (MAB) problem and some of its solutions.
      Despite being a classic RL problem, some of its real-world applications were assessed recently in the literature:
      e.g., online advertising, as demonstrated by \Textcite{PikeASG2018,ChenWYW16};
      resource distribution, studied by \Textcite{ClaureCMJN20};
      and computer networking, shown by \Textcite{VermorelM05}.

   \section{Our Methodology \label{sec:method}} 
      This section is organized as follows.
      \Cref{subsec:model} introduces the proposed approach towards the game of Minesweeper as well as justifies some of our modeling choices.
      \Cref{subsec:msvsmab} displays the differences between the standard MAB problem and Minesweeper, highlighting their implications.
      \Cref{subsec:sim} showcases the considered symmetries and how to obtain them given our modeling of actions.
      \Cref{subsec:TLpot} presents the game characteristics which we believe to be beneficial to transfer learning.

   \subsection{Initial Problem Modeling \label{subsec:model}}
      A Markov Decision Process (MDP) is a mathematical structure which represents a stochastic process.
      It can be described as 4-tuple $(S, A, p, r)$ with components defined as follows:
      \begin{itemize}
         \item $S$ is the set of states of the process;
         \item $A$ is the set of all possible actions, with $A(s) \subseteq A$ being the actions available within state $s \in S$;
         \item $p : S \times S \times A \rightarrow [0, 1]$ is the transition probability function, so that $p(s' | s, a)$ is the probability of reaching state $s'$ from state $s$ as a consequence of action $a$;
         \item $r : S \times A \times S \rightarrow \mathbb R$ is the reward function (also known as signal), so that $r(s, a, s')$ represents the expected reward for taking action $a$ in state $s$ and then reaching state $s'$.
      \end{itemize}

      Straightforwardly, Minesweeper could be modeled as an MDP whose states are whole board configurations, defined by mine coordinates and tile conditions (covered, exposed or flagged).
      In each state unexposed tiles imply possible actions, and there is no uncertainty in the realization of any action:
      $p : S \times S \times A \rightarrow \{0, 1\}$.
      \removed{However, a player is unable to be sure of the state of the game during its course, since mines are hidden.}
      Another aspect to be observed is the fact that there is no benefit in winning the game in as few moves as possible.
      In other words, this modeling would lead to a deterministic, \removed{partially observable,} undiscounted MDP.

      Still in the same regard, a player is unable to be sure of the state of the game during its course, since mines are hidden.
      Thus, this MDP would also be partially observable:
      a belief state would cover all mine arrangements compatible with a given board configuration, and it would be continuously updated as tiles are unveiled.
      Considering such characteristics, this could possibly be approached in a direct fashion using Monte Carlo methods~\cite{KattOA17,SilverV10}.
      However, the computational feasibility of this solution would be questionable, as an example, for games in the intermediate and expert levels, because of the enormous number of states in these conditions.

      Now consider, for the sake of argument, that game states could be perfectly acknowledged, so that it would be possible to compute in a tabular fashion $Q(s, a)$, an estimate of how valuable it is to take action $a$ whenever $s$ is the game state.
      This could in turn be used to always play the best action in any game state.
      In spite of that, such an approach would be computationally intractable even in the beginner level of the game:
      the number of all possible mine arrangements equals to those of 10-combinations of a set with 64 elements;
      each tile can be exposed or not, resulting in $2^{64}$ variations in this sense;
      therefore there would be over $2^{64} \cdot {64 \choose 10} \sim 10^{30} $ states.

      To circumvent this problem, we decided to take inspiration from a human approach of the game:
      looking at numbered tiles around covered ones and probing for those that allow an assertive decision about hiding a mine or not.
      Such strategy capitalizes on the fact that in numerous situations the entire board provides as much information as the neighborhood of a tile under consideration for the aforementioned decision.
      We fulfilled this idea modeling actions as a 10-tuples containing the information of the tiles in a $3 \times 3$ portion of the board plus an action target, which indicates an unexposed tile on the border of the frame that is to be played. 
      \Cref{fig:conf_example} illustrates this concept.
      \begin{figure}
         \centering
            \begin{logicpuzzle}[color=gray!40, rows=7, columns=7, scale=0.5, width=0.4\columnwidth]

               \fillrow{2}{0,1,1,1,1,1}
               \fillrow{3}{0,1,1,1,1,1}
               \fillrow{4}{0,0,1,1,1,1}
               \fillrow{5}{0,0,0,0,1,1}
               \fillrow{6}{0,0,0,0,1,1}

               \setrow{1}{-1,-1,-1,-1,-1,-1,-1}
               \setrow{2}{-1,,,,,,-1}
               \setrow{3}{-1,,,,,,-1}
               \setrow{4}{-1,2,,,,,-1}
               \setrow{5}{-1,1,2,3,,,-1}
               \setrow{6}{-1,,,1,,,-1}
               \setrow{7}{-1,-1,-1,-1,-1,-1,-1}

               \begin{puzzlebackground}
                  \colorarea{blue!40}{\xtikzpath{1}{1}{8/7, 6/1, 2/7, 4/1}}
                  \colorarea{blue!40}{\xtikzpath{7}{1}{8/7, 6/1, 2/7, 4/1}}
                  \colorarea{blue!40}{\xtikzpath{2}{1}{6/5, 8/1, 4/5}}
                  \colorarea{blue!40}{\xtikzpath{2}{7}{6/5, 8/1, 4/5}}
               \end{puzzlebackground}

               \framearea{black}{\xtikzpath{2}{2}{8/5, 6/5, 2/5, 4/5}}
            \end{logicpuzzle}
      
         \caption{
            A framed game board: the blue tiles are out of bounds.
            In this example a mine can be trivially found on coordinate $(3, 2)$.
            This can be inferred based on the 2-tile on $(2, 2)$ and its surroundings, represented by the action $(0, 0, 1, 1, 2, 3, 2, C, C, S)$.
            `C' stands for covered tiles.
            The same is possible based on the 1-tile on $(2, 1)$ considering the action $(-1, 0, 0, -1, 1, 2, -1, 2, C, SE)$.
            The action target (last entry of the tuple) is denoted by cardinal directions abbreviations:
            `S' for South, `SE' for southeast, and so on.
            The other 6 actions which would also target $(3, 2)$ but do not allow to directly detect the mine it features are: 
            $(-1, 1, 2, -1, 2, C, -1, C, C, E)$, $(-1, 2, C, -1, C, C, -1, C, C, NE)$, $(2, C, C, C, C, C, C, C, C, N)$, $(C, C, C, C, C, C, C, C, C, NW)$, $(2, 3, C, \allowbreak C, C, C, C, C, C, W)$ and $(0, 1, C, 2, 3, C, C, C, C, SW)$.
            \label{fig:conf_example}
         }
      \end{figure}

      An upper bound for the number of actions in this setting is $8 \cdot 12^8 \sim 10^{10}$, considering that there are 8 possible action targets and each of the other 8 tiles can be exposed, exhibiting a value from 0 to 8, or be covered and with no flag, or be flagged, or even be outside the game board, totaling 12 cases.
      However, it is important to point that not all of these variations can indeed happen in a Minesweeper game.
      Beyond that, because some of them are symmetric, what is thoroughly discussed in \cref{subsec:sim}, the number of truly distinct actions is considerably below this limit.
      
      Naturally, it could be considered to extrapolate this idea to larger windows.
      However, the expected number of actions appears to quickly become intangible:
      for a $4 \times 4$ window, using the same reasoning as before, there would be up to $12 \cdot 12^{15} \sim 10^{17}$ actions;
      more generally, for a $n \times n$ window, the upper bound on the number of actions would be $4(n-1) 12^{n^2 - 1} \sim 10^{n^2}$.
      It is interesting to point that even if a manageable amount of memory was required for \changed{learning}{memorization} when $n > 3$, covering a variety of actions large enough to play reasonably would still be challenging and unappealing:
      this was confirmed experimentally, as reported in the next section.%

      Another interesting fact regarding this action modeling, relying on neighborhood information only, is that it induces discovering useful game patterns:
      actions which are surely safe or always lead to revealing a mine, regardless of the rest of the board in which one of these actions can be realized.
      An instance of one of such patterns can be found in \cref{fig:conf_example}:
      the action $(-1, 0, 0, -1, 1, 2, -1, 2, C, SE)$, whose target is at $(3,2)$, matches the motif of a 1-tile that neighbors only one covered tile, which consequently must be mine.
      In other words this shifts the focus from $Q(s, a)$ towards $Q(a)$, the action value, taking no account of the board configuration and therefore greatly reducing the computational cost of playing the game.
      This action-only perspective combined with the deterministic and undiscounted attributes of Minesweeper are the cornerstones of its approach as a MAB problem.

      It is interesting to stress that our modeling relies on characteristics that separate Minesweeper from various games, such as Chess:
      for example, any action which does not lead to an immediate loss, while reveals a tile, contributes to winning;
      also, actions are more localized and independent, favoring the use of patterns regarding small sections of the board for playing successfully.
      That inspired the proposed solution, which \removed{is model-free and} takes no account of state transitions or action sequences, as usual in a MAB scenario.

   \subsection{Distinctions From Standard MAB \label{subsec:msvsmab}}

      In a standard MAB setting the goal is to maximize the total reward received from repeatedly choosing from a fixed set of options one action to perform and receiving a corresponding reward next, which is randomly defined according to some hidden stationary probability distribution~\cite{sutton2018reinforcement}.
      Despite some resemblance, this differs in a few ways from the problem presented by Minesweeper, most of all because the set of options from which an action is chosen does not remain unaltered during the course of a game.
      Thus we aimed at adapting existing MAB solutions in order to make the best use of such peculiarities of the game while preserving the guarantees they provide.

      \removed{At first sight} Minesweeper \removed{deviates from the target of reward maximization, as the game} does not have a scoring system, so that the ultimate goal is winning regardless of which or how many actions are performed to do so.
      Thus a possible modeling would be to reward an agent only when a game ends.
      However, finding which actions have the best chance of being safe and which are likely to reveal mines can be related to identifying those that consistently provide good rewards, even if such feedbacks are artificial, designed through Reward Shaping~\cite{ng1999policy, grzes_reward_2017} to guide learning.
      These rewards can be defined according to the only possible action outcomes:
      uncovering of a safe tile, or a mine explosion.
      And since every action which does not immediately terminate the game is good, accomplishing as many of these as possible surely leads to the ultimate objective.
      With all this in mind we established the reward function $r(s, a, s') = 1$ or $-1$ if action $a$ reveals a mine or not, respectively.

      This last reasoning brings attention to the exploration-exploitation dilemma in this context:
      while exploration allows refining the expectation of an action being safe or not, it risks ending the game prematurely due to its sheer nature;
      on the other hand exploitation tends to prolong games, creating opportunities to learn about actions whose occurrence is more frequent in mid- and late-game situations and actually winning.
      Moreover, as the set of actions at each time step can be distinct, what contrasts with MAB defaults, exploration inevitably happens, even when explicitly avoided.

      These characteristics enabled the developed agents to successfully play the game, based on the following action value setup:
      \textit{a priori}, for every action $a$, $Q(a) = -1$ and $N(a) = 0$, where $N(a)$ is the number of times the action was performed;
      whenever $a$ happens, its value is updated as an exponential recency-weighted moving average~\cite[expression (2.3)]{sutton2018reinforcement}, so that $N(a)$ is incremented by 1, and then $Q(a) + \frac{1}{N(a)} (R - Q(a))$ is assigned to $Q(a)$, where $R$ is the reward resulting from the action.
      In this setup every unprecedented action is initialized optimistically~\cite{SzitaL08, Even-DarM01, MachadoSB15}, being considered perfectly safe and, therefore, as valuable as possible as an action.
      This is \removed{specially}significant, most of all from a greedy perspective, as it implicitly enforces some exploration as actions never tried are exploited.
      Moreover, this can be combined with the use of the greater action count $N(a)$ as a tie-breaker when choosing the next action, creating a suitable balance of exploration and exploitation for the agents.
      
      \added{
      With this setup, the $Q$-value of an action represents the average reward received from it.
      This $Q$-value update is essentially a Monte Carlo method~\cite[Section 2.3]{MCTSSurvey} and allows us to find the expected reward from an action.
      When coupled with MAB action selection, this approach closely resembles what is called Flat UCB~\cite[Sections 3.6, 4.1]{MCTSSurvey}.
      }

      The last peculiarity tackled by our modeling presents itself in the form of flags.
      While it is possible to play without using this feature, simply by not clicking known mine tiles, flags offer the opportunity of conveying the information that a tile is probably a mine to other actions, whereas simply not clicking the tile does not.
      This is illustrated in \cref{fig:flag_example}.
      \begin{figure}[ht]
         \centering
         \begin{logicpuzzle}[color=gray!40, rows=5, columns=5, scale=0.5, width=0.296\columnwidth]
            \framepuzzle

            \fillcell{1}{1}
            \fillcell{2}{1}
            \fillcell{3}{1}
            \fillcell{4}{1}
            \fillcell{5}{1}
            \fillcell{1}{2}
            \fillcell{2}{2}
            \fillcell{3}{2}
            \fillcell{5}{2}
            \fillcell{3}{3}
            \fillcell{4}{3}
            \fillcell{5}{3}
            \fillcell{4}{4}
            \fillcell{5}{4}
            \fillcell{4}{5}
            \fillcell{5}{5}

            \setrow{1}{ , , *, *, *}
            \setrow{2}{ , , *, 1, *}
            \setrow{3}{1, 1, F, *, *}
            \setrow{4}{{}, 1, 1}
            \setrow{5}{{}, {} , 1}
         \end{logicpuzzle}
      
         \caption{
            In this hypothetical board state, the tiles marked by an asterisk can be deemed safe due to the flag placed in $(3, 3)$.
            From the point of view of our modeling, the action which ultimately resulted in that flag being placed enabled to deduce that the *-tiles are safe.
            \label{fig:flag_example}
         }
      \end{figure}

      Deciding when to place a flag instead of unveiling a tile is the first question to be answered in order to enable its use.
      Fortunately, the already established reward and action value functions provide the necessary support to such decision:
      instead of always choosing the available action of lowest value to uncover a tile, the action with greatest value can be preferred for flag placement if it has the greatest absolute value.
      This is coherent with the fact that a low action value indicates consistency about the fact that an action is safe just as a high action value with respect to the expectation that an action would reveal a mine.
      Since winning is impossible if a safe tile is flagged, the following directive was used:
      once the number of flags surpasses the number of mines, the flagged tile with the lowest action value is forcibly unveiled.

      Still regarding flags, while uncovering a tile will immediately portray whether this was a good action or not, placing a flag does not provide any instant feedback, what hinders fixing a mistaken negative perception of an action.
      For that reason, in our approach learning also occurs just after the game is finished, regardless of the result being a win or a loss:
      during the game, all actions which ultimately resulted in a flag being placed are recorded;
      when the game ends this record is revisited, checking whether indeed there was a mine where the flags were placed or not, receiving the adequate reward.
      Such delayed learning is only possible thanks to the fact that the mines are revealed when the game is over.
      To take full advantage of such information, all actions which were passed over in the turn when the game was finished also have their values updated as if they were chosen among all others, based on the information that they were hiding a mine or not.

   \subsection{Symmetries \label{subsec:sim}}
      From the perspective of our modeling, we consider 2 distinct cases for symmetries, determined by the tile to be played through an action.
      The first, called the diagonal case, occurs when the tile which is the action target lies in one of the 4 corners of the $3\times 3$ portion of the board the action covers.
      The second, called the cardinal case, takes place when the action target is perpendicularly above, below, or besides the central tile of the $3\times 3$ action configuration.
      The intrinsic differences between these two cases which substantiate the need to treat them separately are illustrated in \cref{fig:sim_example_3}.
      \begin{figure}[ht]
         \centering
         \begin{tabular}{cc}



            \begin{logicpuzzle}[color=gray!40, rows=3, columns=3, scale=0.5, width=0.16\columnwidth]
              \framepuzzle


              \begin{puzzlebackground}
                 \colorarea{gray!40}{\tikzpath{3}{1}{6,8,8,8,4}}
              \end{puzzlebackground}

              \setrow{1}{{}, 1}
              \setrow{2}{{}, 1}
              \setrow{3}{{}, 1}

            \end{logicpuzzle}
            &


            \begin{logicpuzzle}[color=gray!40, rows=3, columns=3, scale=0.5, width=0.16\columnwidth]
              \framepuzzle


              \begin{puzzlebackground}
                 \colorarea{gray!40}{\tikzpath{2}{1}{6,6,8,8,4,2,4}}
              \end{puzzlebackground}

              \setrow{1}{1}
              \setrow{2}{{}, 1}
              \setrow{3}{{}, {}, 1}
            \end{logicpuzzle}
            \\
         \end{tabular}

         \caption{
            This example shows why it is impossible to transform a diagonal case into a cardinal case or vice-versa. 
            The non-center tiles of the action configuration on the left are shifted clockwise to transform it into the opposite case.
            This creates an impossible action configuration, as a 0-tile (blank) cannot neighbor a covered tile.
            \label{fig:sim_example_3}
         }
      \end{figure}

      For the diagonal case, 8 symmetric action configurations are possible, as illustrated by \cref{fig:sim_example_1}.
      As for the cardinal case, another 8 possible symmetric action configurations are also considered, as presented in \cref{fig:sim_example_2}.
      In both cases the first 4 configurations can be obtained by rotating the depicted grids around the center tile by $90^{\circ}$.
      The last 4 configurations are the result of mirroring the first 4 along the axis containing the center tile and the action target.
      \begin{figure}[ht]
         \centering
         \begin{tabular}{cccc}
            \begin{logicpuzzle}[color=orange!40, rows=3, columns=3, scale=0.5, width=0.1846\columnwidth]
               \framepuzzle

               \begin{puzzlebackground}
                  \colorarea{orange!40}{\tikzpath{1}{3}{6,8,4}}
               \end{puzzlebackground}

               \setrow{3}{a, b, c}
               \setrow{2}{h, i, d}
               \setrow{1}{g, f, e}
            \end{logicpuzzle}
            &
            \begin{logicpuzzle}[color=orange!40, rows=3, columns=3, scale=0.5, width=0.1846\columnwidth]
               \framepuzzle

               \begin{puzzlebackground}
                  \colorarea{orange!40}{\tikzpath{3}{3}{6,8,4}}
               \end{puzzlebackground}

               \setrow{3}{g, h, a}
               \setrow{2}{f, i, b}
               \setrow{1}{e, d, c}
            \end{logicpuzzle}
            &
            \begin{logicpuzzle}[color=orange!40, rows=3, columns=3, scale=0.5, width=0.1846\columnwidth]
               \framepuzzle

               \begin{puzzlebackground}
                  \colorarea{orange!40}{\tikzpath{3}{1}{6,8,4}}
               \end{puzzlebackground}

               \setrow{3}{e, f, g}
               \setrow{2}{d, i, h}
               \setrow{1}{c, b, a}
            \end{logicpuzzle}
            &
            \begin{logicpuzzle}[color=orange!40, rows=3, columns=3, scale=0.5, width=0.1846\columnwidth]
               \framepuzzle

               \begin{puzzlebackground}
                  \colorarea{orange!40}{\tikzpath{1}{1}{6,8,4}}
               \end{puzzlebackground}

               \setrow{3}{c, d, e}
               \setrow{2}{b, i, f}
               \setrow{1}{a, h, g}
            \end{logicpuzzle}
            \\
            \begin{logicpuzzle}[color=orange!40, rows=3, columns=3, scale=0.5, width=0.1846\columnwidth]
               \framepuzzle

               \begin{puzzlebackground}
                  \colorarea{orange!40}{\tikzpath{1}{3}{6,8,4}}
               \end{puzzlebackground}

               \setrow{3}{a, h, g}
               \setrow{2}{b, i, f}
               \setrow{1}{c, d, e}
            \end{logicpuzzle}
            &
            \begin{logicpuzzle}[color=orange!40, rows=3, columns=3, scale=0.5, width=0.1846\columnwidth]
               \framepuzzle

               \begin{puzzlebackground}
                  \colorarea{orange!40}{\tikzpath{3}{3}{6,8,4}}
               \end{puzzlebackground}

               \setrow{3}{c, b, a}
               \setrow{2}{d, i, h}
               \setrow{1}{e, f, g}
            \end{logicpuzzle}
            &
            \begin{logicpuzzle}[color=orange!40, rows=3, columns=3, scale=0.5, width=0.1846\columnwidth]
               \framepuzzle

               \begin{puzzlebackground}
                  \colorarea{orange!40}{\tikzpath{3}{1}{6,8,4}}
               \end{puzzlebackground}

               \setrow{3}{e, d, c}
               \setrow{2}{f, i, b}
               \setrow{1}{g, h, a}
            \end{logicpuzzle}
            &
            \begin{logicpuzzle}[color=orange!40, rows=3, columns=3, scale=0.5, width=0.1846\columnwidth]
               \framepuzzle

               \begin{puzzlebackground}
                  \colorarea{orange!40}{\tikzpath{1}{1}{6,8,4}}
               \end{puzzlebackground}

               \setrow{3}{g, f, e}
               \setrow{2}{h, i, d}
               \setrow{1}{a, b, c}
            \end{logicpuzzle}
            \\
         \end{tabular}
      
         \caption{
            Symmetries in the diagonal case.
            The first row exemplifies the action rotations, while the second one shows their respective inverses. 
            The targeted tile in each action indexes has orange color.
            \label{fig:sim_example_1}
         }
      \end{figure}
      
      \begin{figure}[ht]
         \centering
         \begin{tabular}{cccc}
            \begin{logicpuzzle}[color=orange!40, rows=3, columns=3, scale=0.5, width=0.1846\columnwidth]
               \framepuzzle

               \begin{puzzlebackground}
                  \colorarea{orange!40}{\tikzpath{2}{3}{6,8,4}}
               \end{puzzlebackground}

               \setrow{3}{a, b, c}
               \setrow{2}{h, i, d}
               \setrow{1}{g, f, e}
            \end{logicpuzzle}
            &
            \begin{logicpuzzle}[color=orange!40, rows=3, columns=3, scale=0.5, width=0.1846\columnwidth]
               \framepuzzle

               \begin{puzzlebackground}
                  \colorarea{orange!40}{\tikzpath{3}{2}{6,8,4}}
               \end{puzzlebackground}

               \setrow{3}{g, h, a}
               \setrow{2}{f, i, b}
               \setrow{1}{e, d, c}
            \end{logicpuzzle}
            &
            \begin{logicpuzzle}[color=orange!40, rows=3, columns=3, scale=0.5, width=0.1846\columnwidth]
               \framepuzzle

               \begin{puzzlebackground}
                  \colorarea{orange!40}{\tikzpath{2}{1}{6,8,4}}
               \end{puzzlebackground}

               \setrow{3}{e, f, g}
               \setrow{2}{d, i, h}
               \setrow{1}{c, b, a}
            \end{logicpuzzle}
            &
            \begin{logicpuzzle}[color=orange!40, rows=3, columns=3, scale=0.5, width=0.1846\columnwidth]
               \framepuzzle

               \begin{puzzlebackground}
                  \colorarea{orange!40}{\tikzpath{1}{2}{6,8,4}}
               \end{puzzlebackground}

               \setrow{3}{c, d, e}
               \setrow{2}{b, i, f}
               \setrow{1}{a, h, g}
            \end{logicpuzzle}
            \\
            \begin{logicpuzzle}[color=orange!40, rows=3, columns=3, scale=0.5, width=0.1846\columnwidth]
               \framepuzzle

               \begin{puzzlebackground}
                  \colorarea{orange!40}{\tikzpath{2}{3}{6,8,4}}
               \end{puzzlebackground}

               \setrow{3}{c, b, a}
               \setrow{2}{d, i, h}
               \setrow{1}{e, f, g}
            \end{logicpuzzle}
            &
            \begin{logicpuzzle}[color=orange!40, rows=3, columns=3, scale=0.5, width=0.1846\columnwidth]
               \framepuzzle

               \begin{puzzlebackground}
                  \colorarea{orange!40}{\tikzpath{3}{2}{6,8,4}}
               \end{puzzlebackground}

               \setrow{3}{e, d, c}
               \setrow{2}{f, i, b}
               \setrow{1}{g, h, a}
            \end{logicpuzzle}
            &
            \begin{logicpuzzle}[color=orange!40, rows=3, columns=3, scale=0.5, width=0.1846\columnwidth]
               \framepuzzle

               \begin{puzzlebackground}
                  \colorarea{orange!40}{\tikzpath{2}{1}{6,8,4}}
               \end{puzzlebackground}

               \setrow{3}{g, f, e}
               \setrow{2}{h, i, d}
               \setrow{1}{a, b, c}
            \end{logicpuzzle}
            &
            \begin{logicpuzzle}[color=orange!40, rows=3, columns=3, scale=0.5, width=0.1846\columnwidth]
               \framepuzzle

               \begin{puzzlebackground}
                  \colorarea{orange!40}{\tikzpath{1}{2}{6,8,4}}
               \end{puzzlebackground}

               \setrow{3}{a, h, g}
               \setrow{2}{b, i, f}
               \setrow{1}{c, d, e}
            \end{logicpuzzle}
            \\
         \end{tabular}
      
         \caption{
            Symmetries in the cardinal case.
            The first row exemplifies the action rotations, while the second one shows their respective inverses. 
            The targeted tile in each action indexes has orange color.
            \label{fig:sim_example_2}
         }
      \end{figure}

      Taking into account the considered symmetries, the ceiling for the number of action configurations has an eightfold decrease, becoming $12^8 \sim 10^9$.
      While this is indeed a substantial difference, this ceiling still significantly exaggerates how many action configurations can happen in a Minesweeper game.
      For example, it is not possible for an 8-tile to neighbor anything but covered tiles or for a 7-tile to neighbor 2 or more numbered tiles.
      These and many other configurations which are considered in the $12^8$ count would never be found in any real game.

   \subsection{Transfer Learning and Training Optimization \label{subsec:TLpot}}
      Having at least 8 mines is a necessary condition for a game to be able to feature every valid action configuration.
      However, this is not a sufficient condition:
      for example, a 1-tile cannot happen in a degenerate $3 \times 3 \times 8$ game.
      Despite this fact there is no doubt that sufficiently bigger settings allow the occurrence of any action configuration.
      This is just the case for standard Minesweeper levels, which have different board dimensions and number of mines, but the possible actions to be performed by an agent in the beginner level would be the same in the intermediate or expert levels.
      On the other hand, how probable is the presence of an action configuration during a game in each of these settings is not the same.
      This inspires the use of transfer learning:
      to consider cross-settings training for best overall performance. 

      Analyzing the influence of game parameters on learning is the way to look for the ideal training scenario.
      Although the number of mines is one of game attributes which have the most direct influence on learning and playing, how big the board is provides some perspective about this amount.
      Consequently, mine density is the prime factor when accounting for action configuration frequencies, as higher densities tend to produce action configurations with higher numbered tiles.
      This makes most agents trained in low-density settings to perform poorly in high-density games, and vice-versa.
      Moreover, mine density is an indicator of how difficult a game setting is, and its variation affects not only win rate but also the number of plays per game and, ultimately, the number of actions learned in a predefined number of games.

      This last assertion points to another aspect of training optimization, that is \changed{learning}{training} efficiency.
      Straightforwardly, smaller boards generally equate in shorter games while larger boards tend to the contrary.
      And an extremely high or low mine density leads to quick games, as each click has a higher chance of clearing the whole board or exposing a mine.
      At last, a $16 \times 30 \times 99$ game is computationally more expensive to set up and play than a $8 \times 8 \times 10$ one.
      Balancing these points is the key to obtain solid knowledge about a substantial variety of the most frequent action configurations, playing as few games as possible, with board dimensions as small as possible.
      Considering that training is constrained not by the number of games but time, although with a few hindrances, agents trained in settings closer to standard beginner level yielded better results overall than intermediate and expert rivals.


   \section{Experimental Evaluation \label{sec:exp}}
      
      
      Each individual experiment comprised a series of episodes, which were complete Minesweeper games.
      In \cref{subsec:inires} all experiments used a $8 \times 8 \times 10$ setting, the standard beginner level.
      Specific board sizes, and mine amounts or densities are indicated in other subsections.
      For the majority of experiments each agent was trained in $10^6$ episodes and some meaningful statistics were then reported.
      The only exception of this rule were the experiments of \cref{subsec:TLres}, in which the agents were trained during $10^6$ episodes but then tested in varied scenarios for $10^4$ episodes each time.
      The main effectiveness metric used is the win rate, what is coherent with the fact that the game has no scoring system. 
      In this regard is worthy noting that the win rate of an agent steadily increases during its training, tending to what could be seen as a \emph{true} win rate, which can be estimated in tests realized after training win rate is considered stable.
      The experiments were carried out using a workstation with an Intel Core i7-4790K 4.0GHz CPU, 32GB RAM memory, and a GNU/Linux Ubuntu 19 operating system.
      The source code, written considering the 3.6 version of the Python programming language, is publicly available\footnote{\url{https://github.com/CanisKalahari/BanditSweeper}}.
        
      \subsection{Initial Impressions \label{subsec:inires}}
\begin{figure}[t]
            \centering
            \includegraphics[width=\myfigscale\columnwidth]{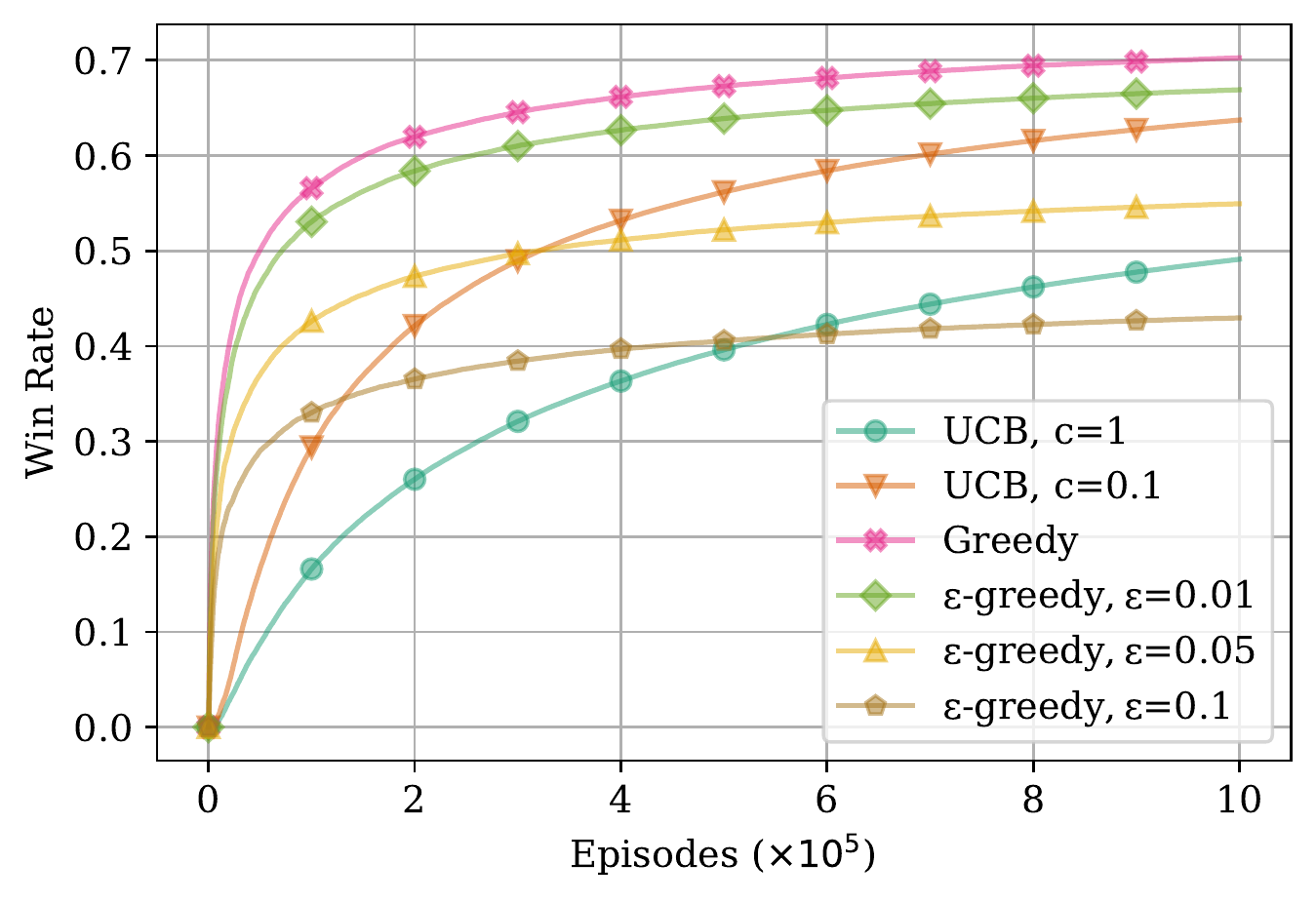}
            \caption{
                  Win rate evolution during training of a variety of agents. 
                  The shape of the curves establishes a clear distinction between UCB and ($\epsilon$-)greedy agents.
               \label{fig:exp_1}
            }
         \end{figure}

         The first experiments targeted to tune MAB algorithms in order to achieve win rates as high as possible, so that the top performing contestants would be further analyzed in other experiments.
         A summary of the results in this preliminary selection can be seen in \cref{fig:exp_1}.
         The highest win rate was that of the strictly greedy agent, followed by an $\epsilon$-greedy one with $\epsilon = 0.01$ and then by tied UCB agents with $c = 0.01$ and $0.1$ (the former was omitted from the graph).
         Although these greedy and $\epsilon$-greedy agents had the highest win rates, the performance of these UCB agents was enough to indicate that exploration is not a problem in itself, and that it could possibly pay off in the long run considering how the curves of the last pair differ from those of the first.
        
\begin{figure}[t]
            \centering
            \includegraphics[width=\myfigscale\columnwidth]{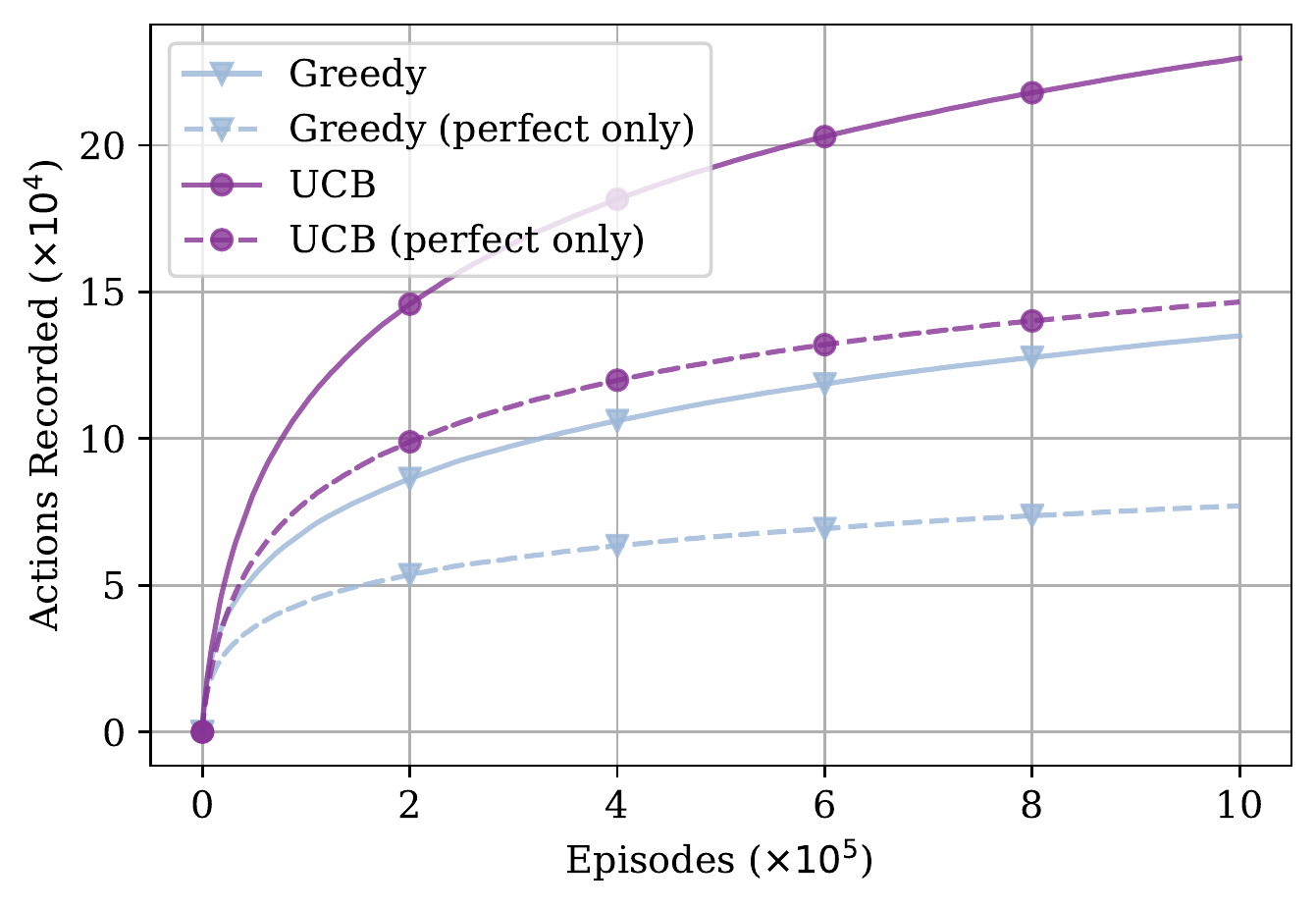}
            \caption{
               Comparison of actions learned over episodes between the two best performing agents from the previous experiment.
               The ``perfect'' tag refers to actions with an action-value of either $-1$ or $1$, so that there is no uncertainty about the result of the concretization of any of them.
               \label{fig:exp_2}
            }
         \end{figure}

         Aside from win rate, another statistic which one could assume to be important in this setting is the number of actions recorded.
         \Cref{fig:exp_2} illustrates this statistic regarding the greedy and UCB ($c = 0.1$) agents.
         While it would be expected for the greedy agent to learn about fewer actions, as its strategy relies exclusively in exploitation, it was first hypothesized that by lasting longer in games it would compensate such trait, coming to possess information about a larger variety of actions.
         Eventually this would lead to getting the number of actions recorded close to that of the UCB agent.
         But it is undeniable that this was not the case:
         despite the lower win rate, the UCB agent recorded almost twice the number of actions the greedy agent registered;
         regarding \emph{perfect} actions (i.e., those with $Q(a) \in \{-1, 1\}$, whose outcome could be fairly expected to be deterministic) the case is similar.
         In short, \changed{learning}{knowing} about more actions does not guarantee to convert such vaster knowledge into higher win rates.

\begin{figure}[t]
            \centering 
            \subfloat{\includegraphics[width=\myfigscale\columnwidth]{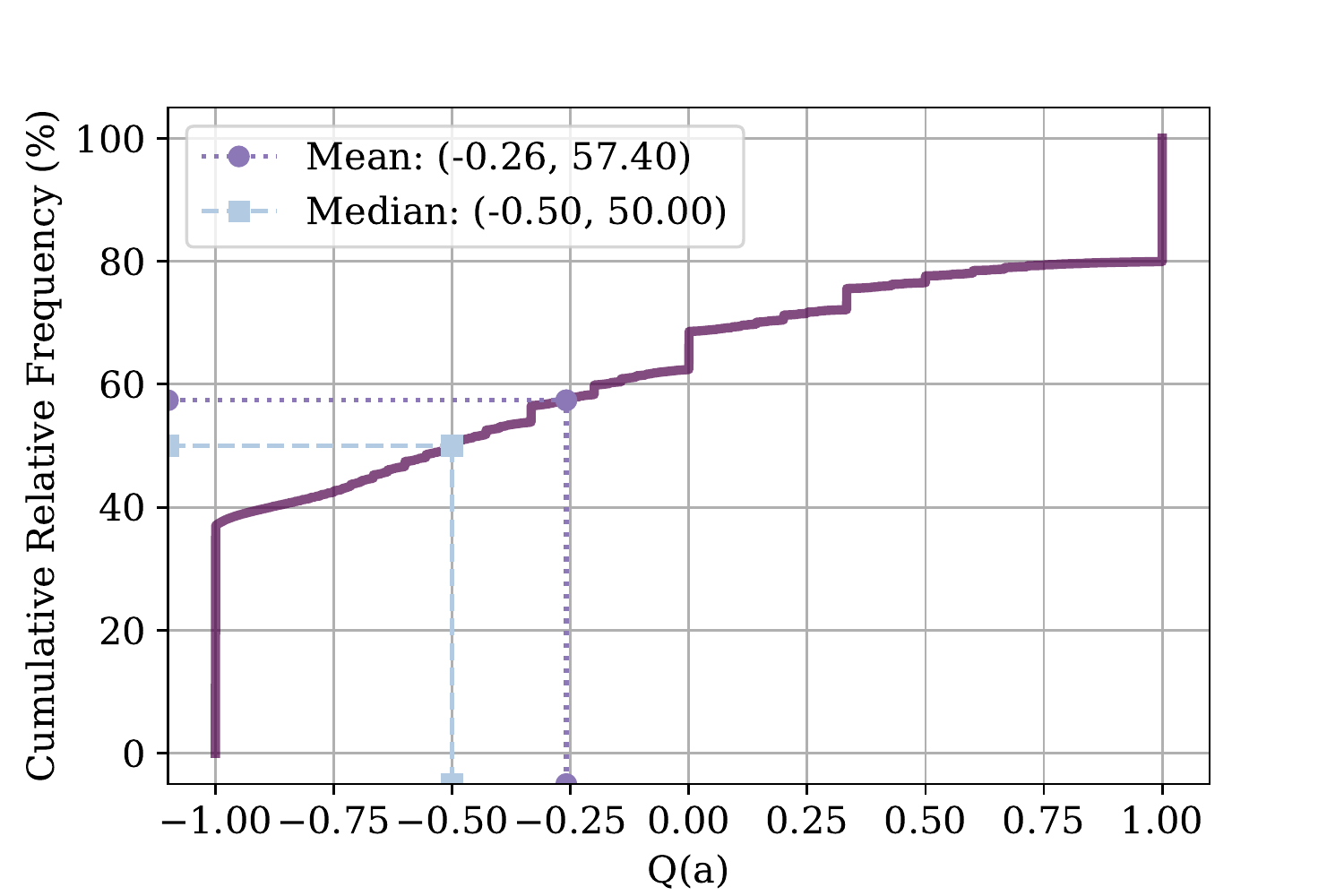}}\\
            \vspace{-1cm}
            \subfloat{\includegraphics[width=\myfigscale\columnwidth]{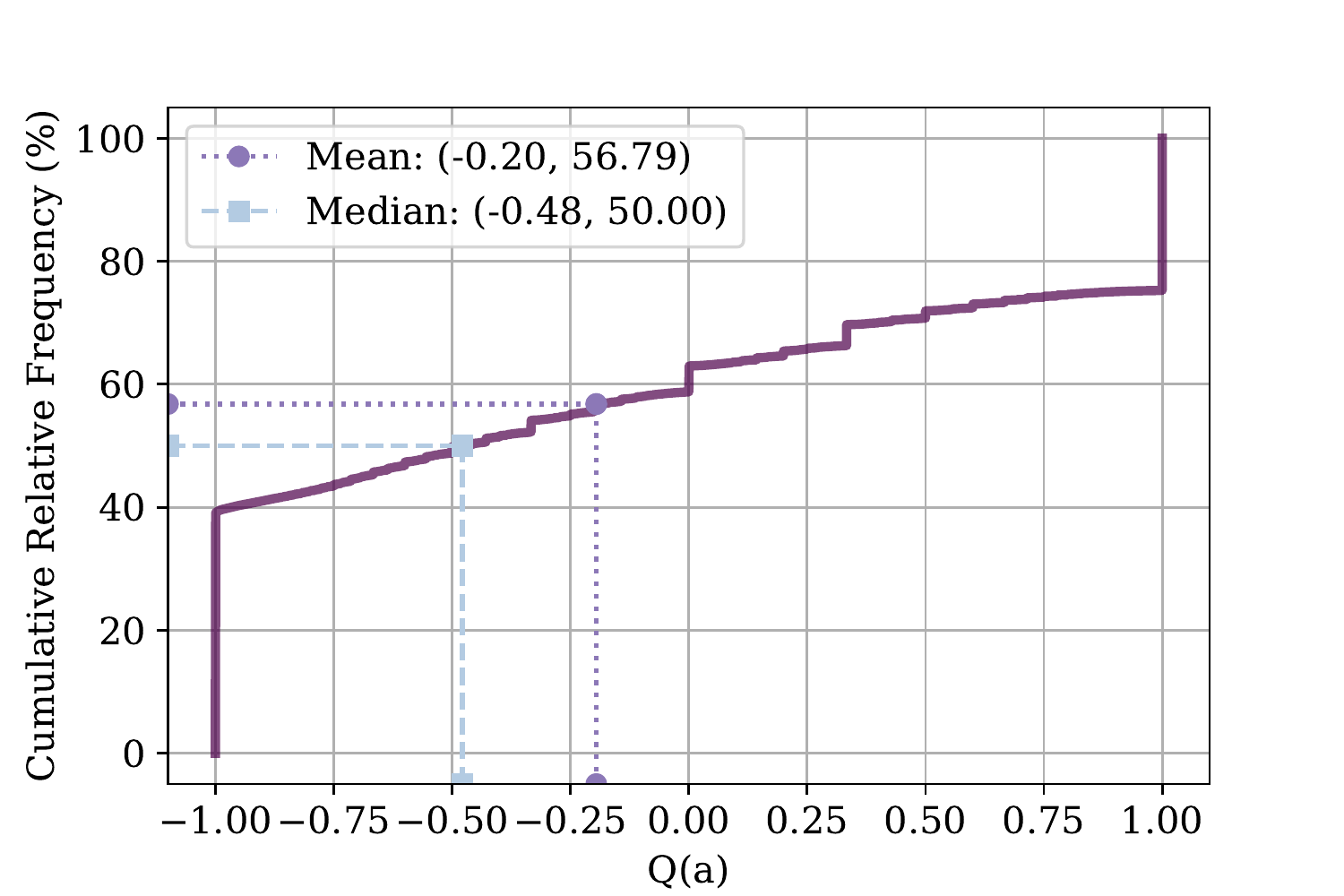}}
            \caption{
               ECDF plots of the action values of the greedy (top) and UCB agents.
               \label{fig:exp_5}
            }
         \end{figure}

         A deeper look at the knowledge accumulated during training could provide an explanation of this phenomenon.
         \Cref{fig:exp_5} presents the plot of the empirical cumulative distribution function (ECDF) of the action values of the selected agents.
         Notably, most actions recorded by both agents are considered perfect, but it should be taken into account that most of these actions have this status only because of having a single past execution, what guarantees such condition.
         The majority of the presumed perfect actions are safe, which is reasonable as it is likely for the number of safe plays needed to win a game to be greater than the number of mines in it.
         On the other hand, the non-perfect actions represent the innate game of chance that Minesweeper can be.
         Overall, both agents have a quite similar profile with respect to $Q(a)$.
         \begin{figure*}[t]
            \centering 
            \resizebox{\myfigscale\textwidth}{!}{\begin{tabular}{rcl}
               \includegraphics{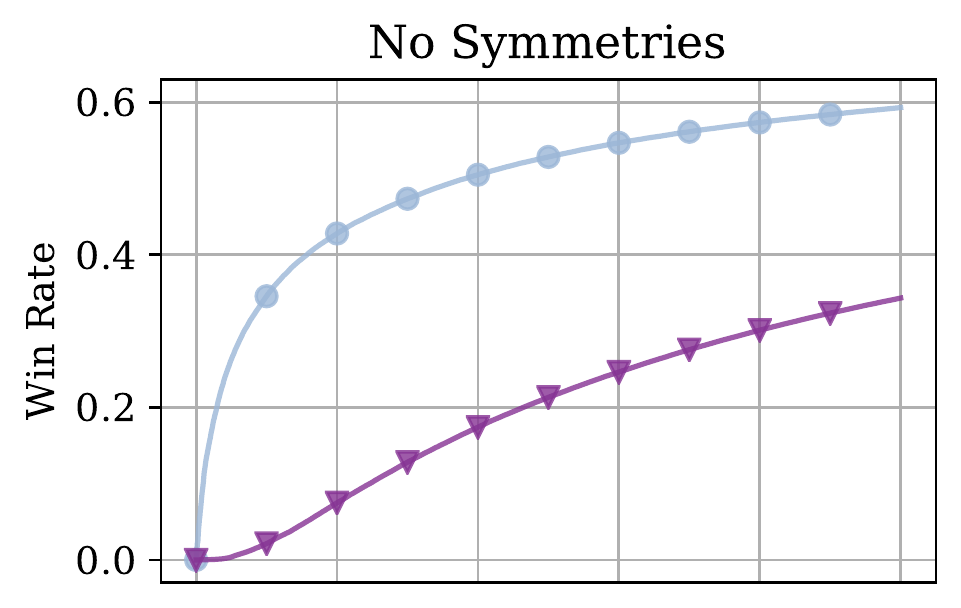} &
               \includegraphics{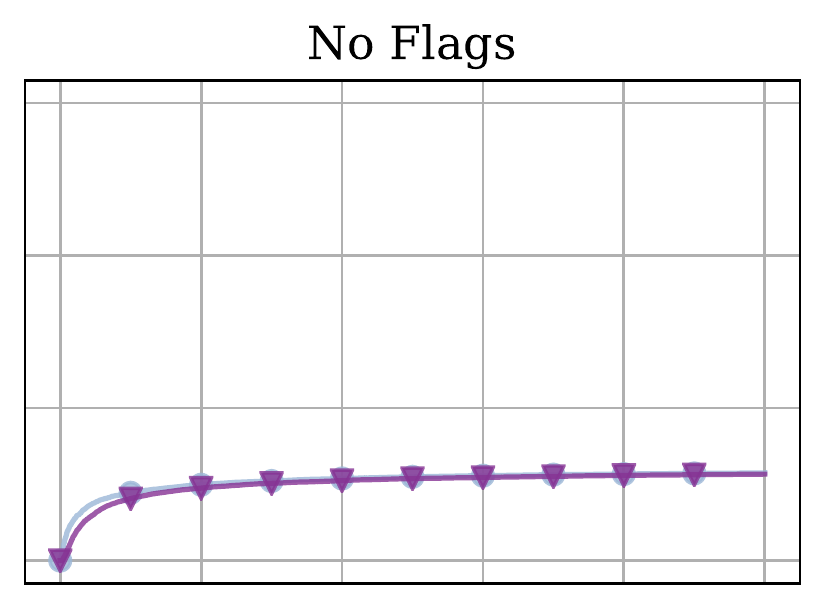} & 
               \includegraphics{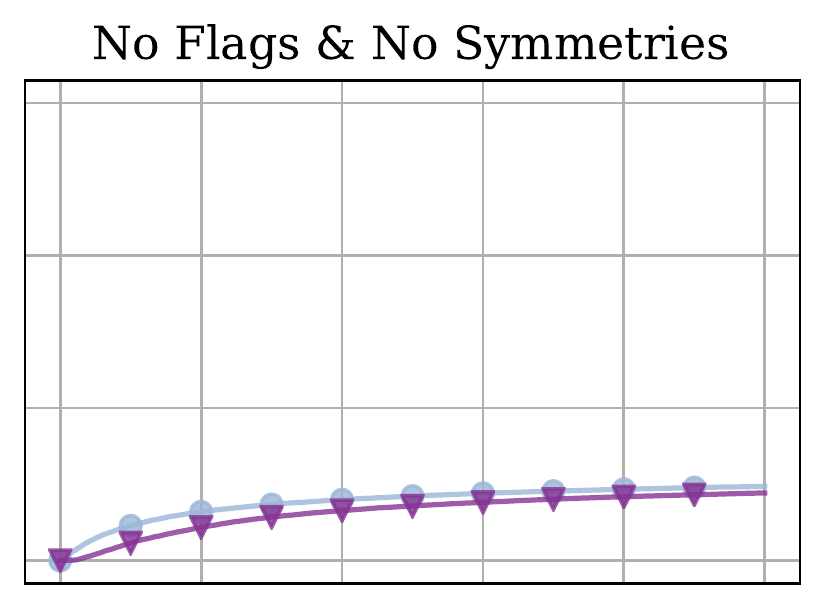} \\
               \includegraphics{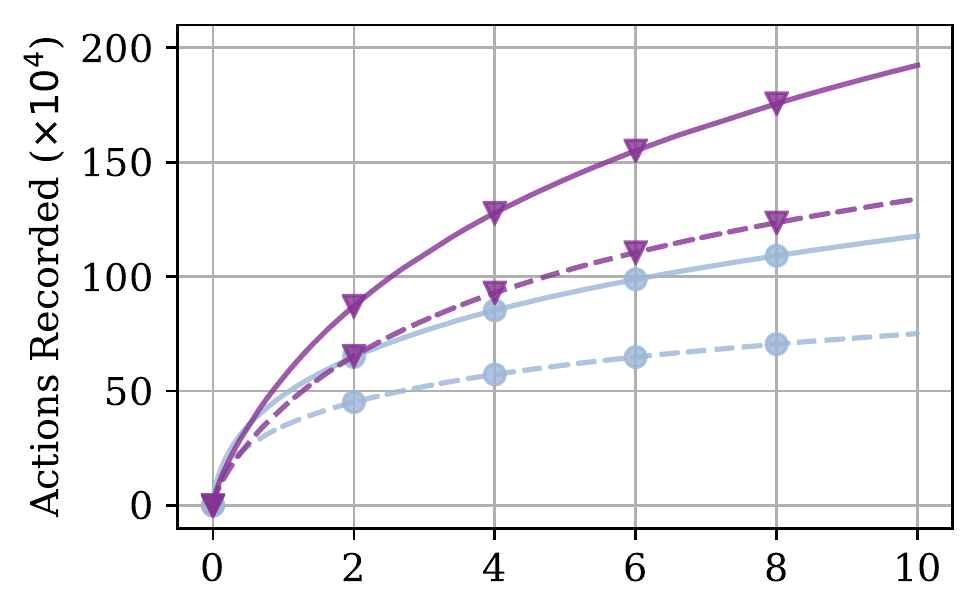} & 
               \includegraphics{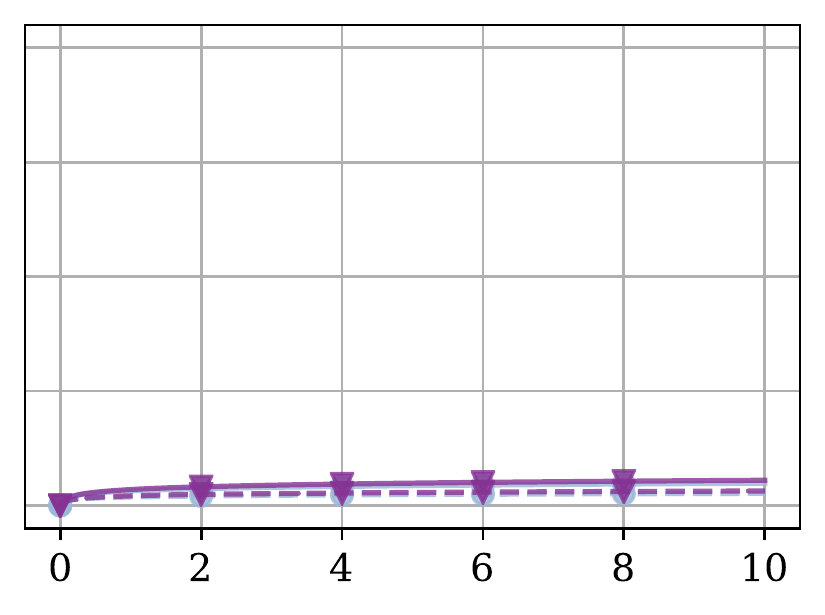} & 
               \includegraphics{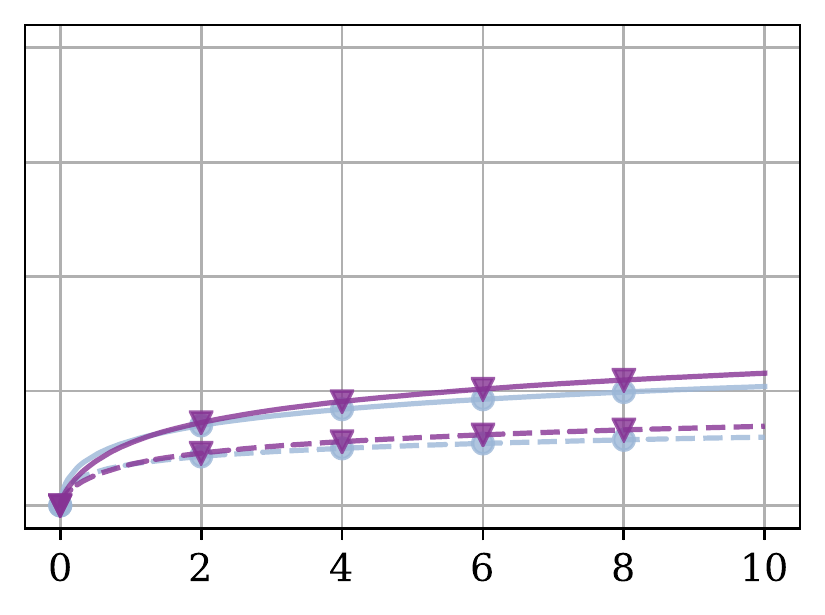} \\
            \end{tabular}}\\
            Episodes ($\times 10^{5}$)

            \vspace{.01\textheight}
            \includegraphics[width=0.7\textwidth]{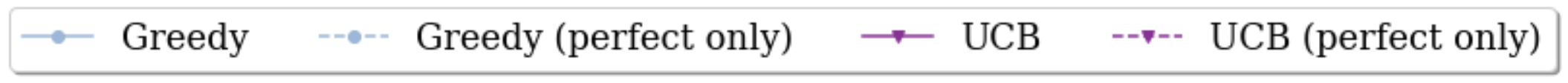}
            \caption{
               Performance of the selected greedy and UCB agents when trained without using flags, considering symmetries, or both.
               \label{fig:exp_4}
            }
         \end{figure*}

         While the distributions of the action values of the selected agents are almost indistinguishable, the action counts provide a more nuanced perspective.
         As shown in \cref{tab:action_cnt}, on average the actions registered by the greedy agent have a greater count than those related to UCB.
         Since the action counts represent how many times the actions were performed, UCB-based schemes experience more variability in the estimation of $Q(a)$~\cite{8969724}.
         Thus it is reasonable to affirm that the greedy agent generally would have a more precise notion of the expected outcome each of the actions it recorded, although such an advantage may disappear in scenarios with higher variability.
         Further, UCB sampling strategy presents the convergence property but it can suffer from unfeasible \changed{learning}{training} time, since the sample mean to converge to the true mean only when the horizon tends to infinity~\cite{7390272,8852326,5425783}.
         At last, exploration is already implicitly carried out by the greedy strategy (see \cref{subsec:msvsmab}), which may help to explain why explicit exploration is not demanded.
            
         \begin{table}[b]
            \centering
            \caption{
               Averages of the action counts of the selected agents
               \label{tab:action_cnt}
            }
            \begin{tabular}{ccc}
               \toprule
               \multirow{2}{*}{Perfect Action?} & \multicolumn{2}{c}{Agent} \\
               \cmidrule{2-3}
               & Greedy & UCB \\
               \midrule
               Yes & \multicolumn{1}{r}{99.2432}  & \multicolumn{1}{r}{51.2799} \\
               No  & \multicolumn{1}{r}{582.6260} & \multicolumn{1}{r}{435.4277} \\
               Both  & \multicolumn{1}{r}{306.7781} & \multicolumn{1}{r}{190.1957} \\
               \bottomrule
            \end{tabular}
         \end{table}

         Still in the same regard, it is noteworthy that the greedy strategy also outperforms the UCB one in the works of~\Textcite{DegrooteHans2016RLfA,IETWang}, with the latter also considering a partially observed Markov decision process problem.
         And based on its superior results compared to those of the UCB agent it could be claimed that the last was farther from finding the balance between knowledge length and depth than its rival.
         In other words, such results reveal that for the considered setting is more advantageous to exploit the arm with the highest expected payoff than to explore poor options motivated by information-gathering purposes~\cite{5308361}.
        
         To assess the impact flags and symmetries have on the selected agents, a collection of experiments was realized in which handicapped versions of the agents were used:
         one just avoided risky tiles instead of explicitly flagging them;
         another considered symmetric actions different, \changed{learning about}{memorizing} each of them separately;
         the third and last one combined the characteristics of the first two.
         The results are portrayed in \cref{fig:exp_4}.
         While disabling symmetries only sets back the progress of the agents, disabling flags seems to greatly hinder their \changed{learning}{performance}, with both agents' win rate reaching less than $12\%$.
         This result emphasizes the importance of placing flags for the proposed approach, showing that even in the beginner level neglecting a detail as such can substantially harm success in the task at hand.
         When symmetries are disabled, as actions which could be seen as equivalent are then considered distinct, there is a substantial increase in the number of actions registered, perfect or not.

   \subsection{Tinkering With Game Mechanics \label{subsec:tinkering}}
      Having scouted the way an agent interacts with a fixed Minesweeper setting, the following experiments were focused on observing the effects game attributes such as board dimensions and mine density have on learning, aiming at finding out how to use them to improve training.
      For this purpose 2 collections of experiments were carried out:
      one varying board width and height from 3 to 10 while approximately maintaining a constant mine density of $0.15625$, which is the same of the standard beginner level;
      and the other with a fixed board width of 10 columns while varying its width from 3 to 10 as well as mine density from $0.14$ to $0.2$.
      These experiments were performed using purely greedy agents, not only because it was the top performer in the initial evaluation but also because its simpler mechanics facilitates realizing the desired analysis.

\begin{figure}[t]
         \centering
         \includegraphics[width=\myfigscale\columnwidth]{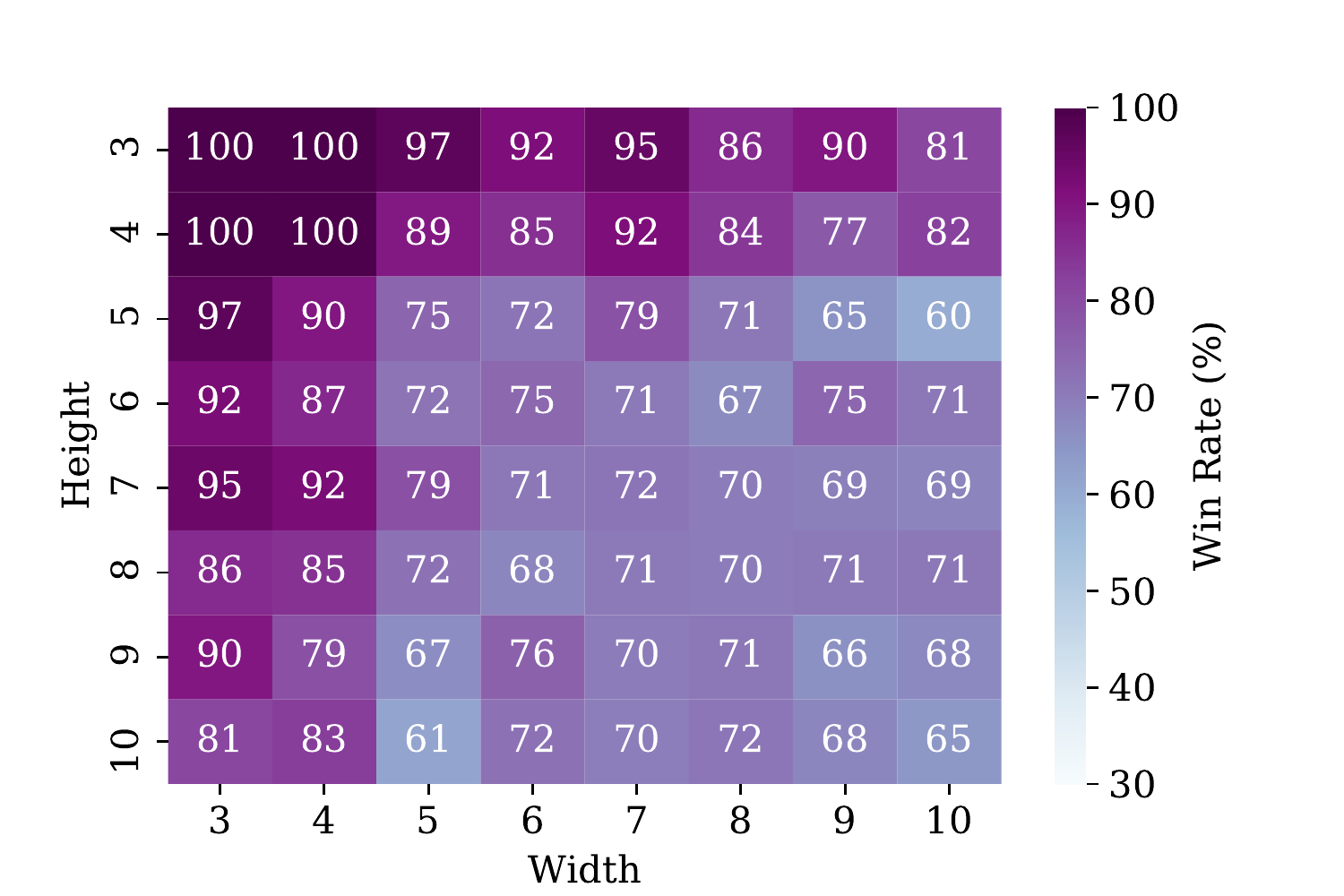}
         \caption{
            Heatmap displaying the win rate of greedy agents trained on varied board dimensions with the number of mines defined rounding the value obtained considering a fixed mine density of $0.15625$.
            Values were rounded to their respective nearest integers, resulting in the $100\%$ win rate cells shown.
            \label{fig:exp_6}
         }
      \end{figure}

      \Cref{fig:exp_6} confirms that board dimensions indeed influence win rate even with a constant mine density.
      This can be explained straightforwardly:
      a larger board would feature a greater number of mines spread across its cells and consequently would probably require a greater number of plays to be defeated;
      as the number of mines increase there is a higher likelihood of risky, non-perfect actions, and eventually they could become the only available options.
      Moreover, it can be noticed that the standard beginner and intermediate levels have the same mine density, but the board of the latter is 4 times bigger than that of the former, so that winning a game of the intermediate level can be compared to winning 4 games of the beginner level in a row.

\begin{figure}[t]
         \centering
         \includegraphics[width=\myfigscale\columnwidth]{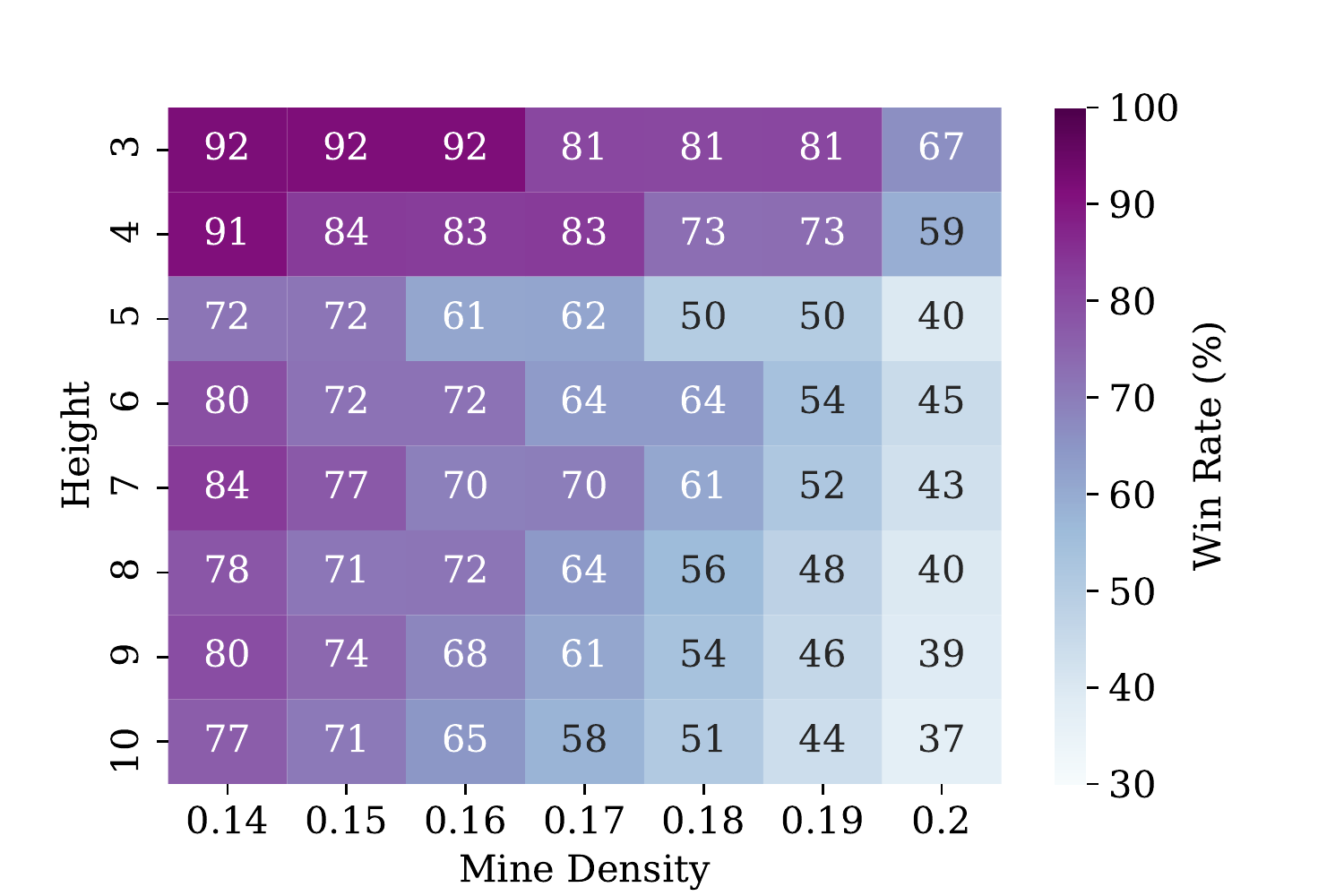}
         \caption{
            Heatmap displaying the win rate of greedy agents trained on varied board heights and mine densities with a fixed width of 10 columns.
            The number of mines is defined rounding the value obtained considering the just mentioned attributes.
            \label{fig:exp_7}
         }
      \end{figure}

      \Cref{fig:exp_7} also presents fairly interpretable results, as by increasing board height the win rate decreases but by increasing mine density the same happens more significantly:
      on average the decrease produced by the first was of $26.1\%$ versus $35.3\%$ of the second.
      Though expected, these results further accentuate the overwhelming importance of mine density when discussing win rate, since an increase by $0.06$ in this attribute could result in a drop in win rate of at least $25\%$, considering the results just reported.
      The combined increase of mine density along with the board dimensions is what makes the expert level brutally harder, as shown and discussed next.

      One last experiment in the same context focused on comparing the proposed MAB-based approach with the one hypothesized in \cref{subsec:model}, in which full game boards would be taken as states of the environment so that learning would happen from a state-action (SA) reference.
      As previously indicated, applying such a method even in the beginner level of the game would be challenging because of the vastness of the learning domain.
      Thus the intended comparison was realized in more favorable settings, considering $4 \times 4$ boards with a varying number of mines.
      \Cref{tab:SAvsMAB} presents the results.
      As the number of mines increase, the same happens to the learning domain as well as to the actions and SA pairs recorded.
      However, such increase is more pronounced for the SA agent, whose performance ceiling is higher but ultimately requires more experience (i.e. games played) in order to reach it.
      \begin{table}[t]
        \centering
        \caption{
           Comparison of MAB and State-Action (SA) greedy agents
           \label{tab:SAvsMAB}
        }
        \begin{tabular}{rrrrrr}
           \toprule
           \mc{1}{c}{\mr{2}{*}{Game Setting}} & \mc{2}{c}{MAB}                                               & & \mc{2}{c}{SA} \\
           \cmidrule{2-3} \cmidrule{5-6}
           & \mc{1}{c}{Win Rate} & \mc{1}{c}{\makecell{Actions\\Recorded}} & & \mc{1}{c}{Win Rate} & \mc{1}{c}{\makecell{SA pairs\\Recorded}} \\
           \midrule
           \mc{1}{l}{$4 \times 4 \times 3$}\\
           $10^5$ games                       &              0.9466 &                                  13940 & &              0.9379 &                15170 \\
           $10^6$ games                       &              0.9583 &                                  14357 & &              0.9612 &                15826 \\
           \midrule                                                                                                                           
           \mc{1}{l}{$4 \times 4 \times 4$}\\                                                                                                 
           $10^5$ games                       &              0.6800 &                                  66573 & &              0.6190 &               117946 \\
           $10^6$ games                       &              0.7462 &                                  82571 & &              0.7462 &               223094 \\
           \midrule                                                                                                                           
           \mc{1}{l}{$4 \times 4 \times 5$}\\                                                                                                 
           $10^5$ games                       &              0.3191 &                                 144988 & &              0.2709 &               233315 \\
           $10^6$ games                       &              0.4229 &                                 247730 & &              0.4137 &               754505 \\
           \bottomrule
        \end{tabular}
      \end{table}

   \subsection{Transfer Learning Results \label{subsec:TLres}}

      Of the game attributes which were covered in the just reported experiments, mine density was the most intriguing, considering how it could be used for training optimization:
      to modify game difficulty while keeping board dimensions unaltered sparked expectations for agents which could be trained in higher density settings to thrive in lower density ones.
      The caveats of such use of mine density are that the win rate is highly sensitive to changes in it, and that if the disparity between the mine density of the board an agent is trained on and the board he is tested on is too large the agent could not be able to successfully employ its knowledge in the test setting. 
      \Cref{tab:exp_1} displays the discrepancy between agents trained with the same board dimensions but with different mine densities.
      By simply adding 3 mines from the standard of 10 mines in the beginner level the training win rate plummets from $70.2\%$ to $41.9\%$ and reached as low as $11\%$ with 17 mines.
      However, noticing that such reduction of training effectiveness does not necessarily produce the same effect on post-training tests, as shown in the bottom 3 rows of the same table, inspired a deeper exploration of such fact.
      \begin{table}[t]
         \centering
         \caption{
            Win rate of selected greedy agents trained and tested in varied scenarios (significantly best in bold)
            \label{tab:exp_1}
         }
         \begin{tabular}{lccccc}
            \toprule
            \mr{2.5}{*}{Game Setting}   & \multicolumn{4}{c}{Scenario}                                                          \\
            \cmidrule{2-5}
                                        &  Training           & Beginner            & Intermediate        & Expert              \\
            \midrule
            $8\times 8 \times 10$       & \boldmath{$0.7025$} & $0.7350$            & $0.4409$            & $0.0028$            \\
            $8\times 8 \times (10..13)$ & $0.5515$            & $0.7520$            & $0.5206$            & $0.0168$            \\
            $8\times 8 \times 13$       & $0.4193$            & $0.7550$            & \boldmath{$0.5707$} & $0.0269$            \\
            $8\times 8 \times 15$       & $0.2504$            & \boldmath{$0.7696$} & \boldmath{$0.5794$} & \boldmath{$0.0413$} \\
            $8\times 8 \times 17$       & $0.1105$            & $0.7570$            & $0.5341$            & $0.0341$            \\
            \bottomrule
         \end{tabular}
      \end{table}

      Still in the same context, it was contemplated the idea of training using not one but a combination of mine densities and along with it arose the question of whether or not it presented any benefits over the more conventional training using a single setting, with the ultimate goal of producing an agent better fit for the intermediate and expert levels.
      Then an agent was made to train in a $8 \times 8$ board with 10 to 13 mines, $25 \cdot 10^4$ episodes for each number of mines, totaling $10^6$ episodes.
      These mine amounts were chosen targeting to cover as close as possible the mine densities of the standard levels from beginner up to expert.
      Though this idea could look promising from start, results displayed in \cref{tab:exp_1} show that an agent subjected to this alternative training routine does not present any true advantages when compared to another one trained in a higher mine density from the beginning.

      Training in higher mine densities can indeed create agents better suited for cross-settings evaluation, but it does not entail that the higher the density the better the agent performs from a transfer learning perspective.
      \Cref{tab:exp_1} also shows that despite the increase in training mine density, there is a decrease in win rate in all standard levels when trained in an setting exaggeratedly dense such as $8\times 8\times 17$.
      These results indicate that mimicking mine density is not enough when creating a successful agent based on transfer learning.
      And as suggested by the results of the $8 \times 8 \times 15$ greedy agent, it can be valid to train in a density higher than the one targeted in order to create a margin for what we suspect to be a form of compensation for the additional difficulty inherent in larger boards.

      \Cref{tab:tl_ucb} presents the results of UCB agents in experiments in the same fashion of those realized with greedy agents.
      More than for the simple sake of completeness, reporting these results is interesting as they show how UCB agents exhibit the exact same behavior of their greedy rivals, profiting from training in a higher density setting for a better overall performance.
      This was enough to make the performance of the $8 \times 8 \times 15$ UCB agent in the beginner level the second best of all tested options.
      Nevertheless, the disparity between the greedy and UCB agents only became more apparent with mine density increase, favoring the first.
      In the end the superiority of the $8 \times 8 \times 15$ greedy agent above all its rivals can be safely stated:
      \removed{%
         every sample of game outcomes (0 for lose, 1 for win) of all experiments can be seen as derived from a normally distributed population according to D'Agostino-Pearson tests with a significance level of 0.01;
         and there is no overlap between respective Student t-test $99\%$ confidence intervals.
      }
      assessing every sample of game outcomes (0 for loss, 1 for win) according to Mann-Whitney tests with a significance level of 0.05 this agent undisputedly prevailed over its rivals in 2 of 3 scenarios, while it was tied for the lead in the remainder.
      \begin{table}[t]
         \centering
         \caption{
            Win rate of selected UCB agents trained and tested in varied scenarios (significantly best in bold)
            \label{tab:tl_ucb}
         }
         \begin{tabular}{lcc}
            \toprule
            \multirow{2}{*}{Scenario} & \multicolumn{2}{c}{Game Setting} \\
            \cmidrule{2-3}
                                      & $8\times 8 \times 10$ & $8\times 8 \times 15 $ \\
            \midrule
            Training                  & \boldmath{$0.6372$}   & $0.1839$               \\
            Beginner                  & $0.7289$              & \boldmath{$0.7644$}    \\
            Intermediate              & $0.2699$              & \boldmath{$0.4453$}    \\
            Expert                    & $0.0000$              & \boldmath{$0.0062$}    \\
            \bottomrule
         \end{tabular}
      \end{table}
  
   \section{Conclusion \label{sec:end}}
      Minesweeper provides a challenging task for any agent to master.
      Past works with this goal mostly relied on CSP and heuristic approaches, which could be considered incontrovertible choices taking into account some game aspects.
      Ultimately they could provide very good win rates, the most basic success benchmark in this context.
      However, this subject appeared to be far from being exhausted.
      We suspected that if the problem modeling contemplated some other key game properties, then RL could be fruitful in this setting, not only in the sense of winning but also for effectively investigating the game from a novel perspective.
      This motivated the development of the proposed MDP modeling as well as the adaptations of classic MAB algorithms $\epsilon$-greedy and UCB.

      The best performing agent resulting from our methodology was able to achieve a win rate of over $70\%$ while learning Minesweeper from scratch in $10^6$ games of the standard beginner level.
      Trying to put this in context, this could be compared to the best 2.5\% results of human players~\cite{world_minesweeper_best}.
      Providing such a detailed description of the conditions to which the agent was subjected, instead of just the win rate, is uttermost important since we are more interested in learning than in winning.
      The same goes for the fact that the just mentioned agent is purely greedy, and that it outdid its rivals despite learning in significantly more succinct fashion.
      Moreover, we sought to optimize training inspired in transfer learning, what resulted in the just mentioned best agent obtaining win rates of $76.96\%$, $57.94\%$ and $4.13\%$ in the beginner, intermediate and expert levels, respectively.

      The main obstacle we found while experimenting with Minesweeper was the pure brutality of large boards' difficulty when tackled by our approach.
      Training over $10^6$ games in a $8 \times 8 \times 15$ setting took $14$ hours, while playing $10^4$ $16 \times 30 \times 99$ games took almost $7.5$ hours.
      This makes it virtually impossible to train on larger boards, which led us to use transfer learning as an alternative.
      But still, while results on the intermediate level are satisfactory, the same cannot be said about the expert level.
      
      In future works we hope to perfect the balance of learning and game settings to define an agent capable of handling larger boards to an acceptable degree.
      Another possible direction is one of investigating the use by human players of $3 \times 3$ game patterns discovered by computational agents, what could be set in the context of explainable artificial intelligence.

   \section*{Acknowledgments}
      Igor Q. Lordeiro (Scientific Initiation Scolarship/Bolsista PIBIC) would like to thank CEFET-RJ for financial support. The authors thanks the support of CAPES (code 001), FAPERJ and CNPq.

   \ifCLASSOPTIONcaptionsoff
      \newpage
   \fi




   \renewcommand*{\bibfont}{\footnotesize}
   \printbibliography
\end{document}